\documentclass[10pt,twocolumn,letterpaper]{article}

\usepackage[pagenumbers]{cvpr}      
\definecolor{cvprblue}{rgb}{0.21,0.49,0.74}
\usepackage[pagebackref,breaklinks,colorlinks,allcolors=cvprblue]{hyperref}
\usepackage{hyperref}
\usepackage{url}
\usepackage{algorithm}
\usepackage{algorithmic}
\usepackage{arydshln}
\usepackage{graphicx}
\usepackage{multirow}
\usepackage{float}

\def\paperID{} 
\def\confName{CVPR}
\def\confYear{2026}

\title{Motivation Is Something You Need}

\author{Mehdi Acheli\\
SAMOVAR, Telecom SudParis\\
Institut Polytechnique de Paris, Palaiseau, France\\
{\tt\small mehdi.acheli@telecom-sudparis.eu}
\and
Walid Gaaloul\\
SAMOVAR, Telecom SudParis\\
Institut Polytechnique de Paris, Palaiseau, France\\
{\tt\small walid.gaaloul@telecom-sudparis.eu}
}
\begin{document}
\maketitle
\begin{abstract}
This work introduces a novel training paradigm that draws from affective neuroscience. Inspired by the interplay of emotions and cognition in the human brain and more specifically the SEEKING motivational state, we design a dual-model framework where a smaller base model is trained continuously, while a larger motivated model is activated intermittently during predefined "motivation conditions". The framework mimics the emotional state of high curiosity and anticipation of reward in which broader brain regions are recruited to enhance cognitive performance. Exploiting scalable architectures where larger models extend smaller ones, our method enables shared weight updates and selective expansion of network capacity during noteworthy training steps. Empirical evaluation on the image classification task demonstrates that, not only does the alternating training scheme efficiently and effectively enhance the base model compared to a traditional scheme, in some cases, the motivational model also surpasses its standalone counterpart despite seeing less data per epoch. This opens the possibility of simultaneously training two models tailored to different deployment constraints with competitive or superior performance while keeping training cost lower than when training the larger model.
\end{abstract}    
\section{Introduction}
Modern deep learning is mainly inspired by neurobiology and cognitive science. Examples include the basic multi-layer perceptron architecture, RNNs inspired by feedback loops in the cerebellum \cite{grossberg2013recurrent} and the experience replay in deep Q‑learning (DQN) and how it relates to the hippocampus replaying experiences during sleep \cite{hassabis2017neuroscience}. One of the latest developments that spurred an AI revolution is the transformers architecture. It is also based on an important cognitive process: attention \cite{vaswani2017attention}. Through the integration of this mechanism, neural networks are able to attach different weights or degrees of importance to inputs and features. They adapt these weights dynamically during training to best achieve a downstream task often modeled by a loss function. Transformers have been successfully applied and advanced the state-of-the-art in a wide array of applications: NLP \cite{brown2020language}, Computer Vision \cite{dosovitskiy2020image}, Audio and Speech Recognition \cite{baevski2020wav2vec}, etc. In turn, many theories were proposed as to why humans evolved emotions. They can be seen as social signals \cite{weisfeld2013applying} or as preparatory mechanisms preceding adaptive actions \cite{keltner1999social} with most research focusing on their social utility \cite{van2022social}. However, emotions have a deep impact on learning and memory too \cite{johnson2024effect, tyng2017influences}. While positive emotions bolster effective learning and support long-term memory formation, negative affects like fear and stress impair both. Extreme negative emotions in disorders like depression are associated with severe cognitive dysfunction \cite{johnson2024effect}. 

One particular framework classifies emotions into seven primary states: SEEKING, RAGE, FEAR, LUST, CARE, PANIC/GRIEF, and PLAY \cite{panksepp2004affective}. SEEKING is the one we focus on in our study. Termed "appetitive motivational system", it is associated with a heightened state of curiosity, exploratory tendencies and motivated behaviors powered by the anticipation of reward. This motivated state has been shown to recruit or activate bigger and wider brain regions, ultimately improving cognitive performance \cite{engelmann2009combined}.

Our goal is to replicate this mechanism during the training of artificial neural networks. We propose a task-agnostic framework composed of four elements: the base model, the motivated model, the motivation condition and the weights map. In order to validate this framework, we instantiate it through the image classification task. In essence, during training, we train what we call a base model and we mimic the motivated state at the occurrence of events of interest. We call those events, the motivation condition. For example, if we consider the image classification task, the motivated state would correspond to moments when the network consistently reduces its loss for several batches, analogous to a human learner feeling rewarded by understanding a concept. Once in this state, a bigger neural network, which we call motivated model, is trained until the motivation condition becomes invalid. At this point, we switch back to training the base model. We choose scalable neural networks (ResNet~\cite{he2016deep}, EfficientNet~\cite{tan2019efficientnet}, ViT~\cite{sharir2021image}, etc) to exemplify our theory because they naturally lend themselves to our framework. In most cases, scalable neural networks are a sequence of models where each one is deeper or wider than the previous one. Bigger models are extensions of smaller ones with more layers or more parameters. This simulates how in the brain, bigger networks contain smaller ones. The motivated model is thus a bigger model in the scalable architecture sequence.

\begin{figure*}
    \centering
    \includegraphics[width=1\linewidth]{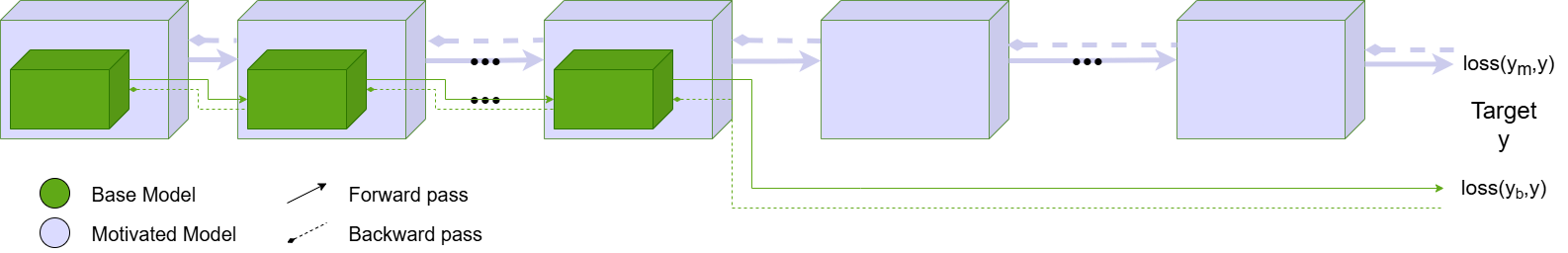}
    \caption{Overview of the interplay between the two networks in our motivation-inspired dual training procedure}
    \label{fig:method}
\end{figure*}

\Cref{fig:method} shows how the two networks are connected. Essentially, we are training one network. We train the base part of it until motivation occurs and then we train it whole until the end of the motivated state. This means that the weights of the base model are continuously updated during the whole training while the weights specific to the motivated model, which we call differential layers, are updated only during the motivated state. The weights map component of our framework, defines where the base model is placed inside the motivated model. For example, the blocks of the base model could correspond to the first blocks of the motivated model as in \cref{fig:method} as well as to later blocks.

For this work, our contributions are threefold: (i) A neuroscience-inspired dual-model training framework that outputs two models and alternates between a base and a motivated model to emulate motivational states during learning. (ii) An instantiation of this framework for the image classification task using scalable architectures (ResNet, EfficientNet and ViT). (iii) An extensive experimental validation that shows how: (1) We efficiently improve the performance of the base model compared to its standalone training. As such, we create intermediary models in the scalable architecture with respect to the training FLOPs count while keeping inference cost the same, helping teams with limited resources.  (2) For some configurations involving EfficientNet, we significantly and efficiently improve the performance of the motivated model while keeping comparable performance of the base model with respect to the standalone counterparts. The dual training cost is lower than the standalone motivated model training. This allows for a train once, deploy twice scheme where  different  deployment targets exist, each with its own resource constraints.

Consequently, in the same way the conditional activation of the broader brain regions in humans has a power-saving effect, our framework also avoids training a full motivated model from start to finish. The analogy meets some limits, however, in the context of offline learning as artificial neural networks undergo training and inference as two separate procedures. In the artificial context, our training paradigm preserves the power-saving feature during the training but also improves the base model performance during the inference step while the cost of the inference stays the same. It creates an intermediary model between the standalone base network and the standalone motivated network, but only training-wise. The inference costs are kept the same for both models. This is increasingly important in today's AI landscape where computational efficiency is closely examined due to its economic and ecological implications.

\section{Related work}
\label{sec:relwork}
From the perspective of the motivated model, our method falls within a broad body of research that acts on parameters participation during training. These include Dropout~\cite{srivastava2014dropout} which greatly improves overfitting by randomly dropping units and Stochastic Depth~\cite{huang2016deep} where entire layers are dropped randomly for certain minibatches in networks equipped with skip connections. Our method can be seen as dropping the differential layers, although not randomly but at specific points defined by the motivation condition. Other works focus on freezing layers, meaning, to stop updating their weights. While ForwardThinking \cite{hettinger2017forward} trains one layer at a time and freezes previously trained layers, FreezeOut \cite{brock2017freezeout} freezes the early layers of the network by annealing their learning rates. The authors have indeed argued that the first layers converge more quickly than later layers. \cite{xiao2019fast} introduce a metric called Freezing Rate based on the gradients of the weights to intelligently choose the layers to freeze. Layers where weights are updated in both positive and negative directions leading to the updates cancelling each other can be frozen. Compared to the first works, this method does not keep the layers frozen the whole training and unfreezes them at fixed intervals. Our algorithm also does not deactivate the motivated model permanently. \cite{cruciata2024learn} reuses the idea of \cite{xiao2019fast} but drops layers instead of freezing them reducing not only backward but also forward computation. The framework Egeria \cite{wang2023egeria} relies on an assistant model to assess the progress of internal layers training and freeze those that converged. This saves computation and communication time during the backward step.

Enforcing weight sparsity is another way to limit parameter participation during training. \cite{dettmers2019sparse} introduce Sparse Momentum to dynamically prune and grow weights using averaged gradient signals. \cite{liu2020dynamic} use trainable pruning thresholds within a unified optimization process. In \cite{curci2021truly}, the authors provide a parallel training algorithm and a true implementation of sparse matrices calculations in a novel software framework. The new framework goes beyond the simulated sparsity in popular deep learning frameworks. Works focusing on freezing or sparsity aim for computational efficiency while our training recipe improves performance as well.

When we consider our whole training procedure, it may be reminiscent of Multi-Task Learning (MTL) \cite{crawshaw2020multi} in which multiple machine learning models are trained for multiple tasks using shared representations. In particular, the way the base model is shared echoes Shared Trunk architectures \cite{crawshaw2020multi}. However, contrary to MTL, we train for a single task using a unique dataset. Besides, separate backward passes per model are not the standard in MTL. Instead, a total loss is considered as a weighted average of task-specific losses and is back-propagated. Some works do employ independent backward passes. \cite{baytas2016asynchronous} apply task-specific gradient steps in an effort to accelerate training when different tasks are mapped to distinct nodes in distributed settings. \cite{lee2022multitask} decouple gradient updates of task-specific layers from those of shared layers creating alternating backward passes. The rationale being that such updates allow the layers to better match their roles as task specific or generic layers.

Our motivation-inspired training also exhibits conditional computation \cite{scardapane2024conditional}. However, conventionally, this concept is associated with networks that dynamically activate or de-activate parts of their computation graph conditionally on their inputs \cite{scardapane2024conditional}, not on training-related information like in our case. Moreover, for our case, there is no conditional activation at inference time like in Early Exit Neural Networks (EENNs)~\cite{teerapittayanon2016branchynet} or Mixture of Experts Networks (MoEs)~\cite{riquelme2021scaling}. The whole base or motivated model is used. In contrast, EENNs involve all the network during training~\cite{teerapittayanon2016branchynet} while some works in MoEs additionally train a routing policy through reinforcement learning~\cite{clark2022unified, rosenbaum2017routing}.  
\section{Proposed approach}
This section presents the four components of our training recipe: the base model, the motivated model, the weights map and the motivation condition as well as the training algorithm. Our method can be seen as a general framework and is domain-agnostic. Hereafter and in the experiments, we instantiate it with the image classification task from Computer Vision.
\subsection{The base model}
The base model can be any neural network. It is trained whole for each batch and each epoch. Once motivation occurs, its weights and buffers are copied into the corresponding part of the motivated model as defined by the weights map and training is resumed on the bigger model.
\subsection{The motivated model}
The motivated model is a network that has more neurons than the base one. It can be deeper (more layers), wider (more neurons per layer) or both. When the motivated model is trained, the base model that it contains is trained simultaneously, as the weights and buffers of that internal model are updated through backpropagation. In practice, those internal weights and buffers are copied to the base model following the weights map once the motivated state stops.
\subsection{Weights map}
Since the motivated model can contain the base model in many different ways, a weights map that associates the exact weights in the base model with their equivalents in the motivated model is needed. Let's take the example of a convolutional layer. In many scalable architectures like EfficientNet, convolutional modules of larger models have a bigger output dimension. With respect to the motivated convolutional layer of output dimension $d'_{out}$, the base convolutional layer of dimension $d_{out}$ can be extracted by combining any $d_{out}$ outputs among the $d'_{out}$ possible ones. The number of combinations is $\binom{d'_{out}}{d_{out}}$. In the following three sections, we explain how the weights map was defined for the three scalable architectures we experimented on.

\subsubsection{Resnet}
Resnet architectures we leveraged for our experiments (ResNet-50, ResNet-101 for ImageNet and ResNet-20, ResNet-32, ResNet-44, ResNet-56 for CIFAR-10 and CIFAR-100) have identical layer types inside each block across architectures. The number of blocks is also the same. The only difference is the number of layers, i.e., the depth of the network. For a block of length $l$ in the base model and $l'$ in the motivated model, we map the last layer of the base block to the last layer of the motivated block while we map the remaining layers in the base model to the first $l - 1$ layers in the motivated one. Other than the blocks, the convolution stem and the classifier head are identical across architectures and mapped together.

\subsubsection{ViT}
ViT-Tiny and ViT-Small networks have the same number of layers. Inside each layer, they differ in the embedding dimension and the number of attention heads. Implementation-wise, however, their weights all belong to simple Linear or LayerNorm layers. LayerNorm weights and all biases are one-dimensional. For a weight of size $d_{out}$ in the base model and $d'_{out}$ in the motivated one, we associate the base model to the first $d_{out}$ scalars in the motivated model. As for the weights of Linear layers, they are two-dimensional. We map the $d_{in}$ input features of the base layer to the first $d_{in}$ among the $d'_{in}$ input features of the motivated layer. We also map the $d_{out}$ output features of the base layer to the first $d_{out}$ among the $d'_{out}$ output features of the motivated layer. ViT architectures have two other special weights that are three-dimensional: the classification token and the positional embedding. These have the same shape across architectures except the last dimension, of size $d$ in the base model and $d'$ in the motivated one. We map the base weight to the first $d$ among $d'$ channels along the last dimension in the motivated weight.

\subsubsection{EfficientNet}
For a weight matrix of one dimension (vector), sized $d_{out}$ in the base model and $d'_{out}$ in the motivated one, we associate the base model to the first $d_{out}$ scalars in the motivated model. This applies in particular to the weights and buffers of Batch Normalization layers and biases of all layers. We expand the same reasoning to weight matrices of convolutional layers. These include four dimensions, the two last ones being the same across architectures since the same kernel is used. The first two dimensions are the input and output channels of the convolutional layers. We associate the $d_{in}$ input channels in the base layer to the first $d_{in}$ of the $d'_{in}$ channels in the motivated layer. Similarly, we associate the output channels $d_{out}$ in the base layer to the first $d_{out}$ of the $d'_{out}$ channels in the motivated layer. If a stage has $l$ layers in the base model and $l'$ layers in the motivated model, we associate the weights of the base model to the first $l$ layers in the motivated one.

\subsection{The Motivation condition}
\label{sec:motivcond}
The motivation condition is the condition which, upon satisfaction, leads to the training of the motivated model instead of the base one. It represents the trigger of a state of curiosity and a high anticipated reward. One can imagine a student during a learning session. If they understand the premise of a subject, they get motivated and more inclined to explore it deeper recruiting new circuitry in their brain. During the training of our image classification model, continued correct classification creates an analogous state. We can thus choose to trigger motivation if the loss decreases for k consecutive batches.

\subsection{Training algorithm}
\Cref{alg:motivated_train} represents a high level pseudo-code for our alternating training procedure. A specific variable $motivated$ holds the state of the training and according to its value, the base model or the motivated model is activated in the forward and backward passes (\textbf{line 1, 7 and 14}). When the loss decreases for $k$ consecutive batches, we update the $motivated$ variable (\textbf{line 24}). Upon the switch, i.e. when the latter variable changes values (\textbf{line 25 and 31}), we copy the weights from the currently active model to the new one. There is an automatic switch in-between epochs if the last epoch ended on a motivated state (\textbf{line 39}) since each epoch starts on a non motivated state. The functions copying weights are architecture-specific and must be customized in accordance with the weights map. Moreover, depending on the optimizer used during training, a "state" is associated with the optimizer which contains buffers for each weight. These buffers are involved in weight updates and integrated with gradient information. Specific functions are thus dedicated to copying these buffers, also following the weights map.

\begin{algorithm}[!htbp]
  \caption{Motivation‑Aware Model Training}\label{alg:motivated_train}
  \begin{algorithmic}[1]
    \STATE motivated $\gets$ \textbf{false};
    
    \STATE Initialize base\_model, motivated\_model, optimizers
    \STATE prev\_loss $\gets$ $+\infty$ 
    \STATE consecutive\_improve $\gets 0$
    
    \FOR{epoch = 1 to $N$}
      \FOR{each inputs, targets in batches}
        \IF{not motivated}
          \STATE outputs $\gets$ base\_model(inputs)
        \ELSE
          \STATE outputs $\gets$ motivated\_model(inputs)
        \ENDIF

        \STATE loss $\gets$ ComputeLoss(outputs, targets)
        \STATE Backpropagate(loss)
        \IF{not motivated}
          \STATE optimizer.step()
        \ELSE
          \STATE optimizer\_motivated.step()
        \ENDIF
        \IF{loss decreased compared to prev\_loss}
          \STATE consecutive\_improve += 1
        \ELSE
          \STATE consecutive\_improve = 0
        \ENDIF

        \IF{consecutive\_improve $\ge$ k}
          \IF{not motivated}
            \STATE motivated = \textbf{true}
            \STATE copy\_weights\_small\_big(base\_model, motivated\_model)
            \STATE
            copy\_state\_small\_big(optimizer, optimizer\_motivated)
          \ENDIF
        \ELSE
          \IF{motivated}
            \STATE motivated = \textbf{false}
            \STATE copy\_weights\_big\_small(base\_model, motivated\_model)
            \STATE
            copy\_state\_big\_small(optimizer, optimizer\_motivated)
          \ENDIF
        \ENDIF

        \STATE prev\_loss $\gets$ loss
      \ENDFOR
      
      \IF{motivated}
            \STATE motivated = \textbf{false}
            \STATE copy\_weights\_big\_small(base\_model, motivated\_model)
            \STATE
            copy\_state\_big\_small(optimizer, optimizer\_motivated)
      \ENDIF
    \ENDFOR
  \end{algorithmic}
\end{algorithm}

\section{Experiments}
\label{sec:exp}
We implemented our training procedure using Pytorch \cite{paszke2019pytorch} and timm \cite{wightman2019pytorch} and conducted our experiments on servers with NVIDIA V100 32G or NVIDIA A6000 GPUs. Code will be made publicly available and is included in the supplementary material. As stated earlier, we evaluate our method on Image Classification for Computer Vision. We consider ResNet architectures for CIFAR-10, CIFAR-100 \cite{krizhevsky2009learning} and ImageNet \cite{deng2009imagenet} in addition to the ViT architecture on both CIFAR datasets and the EfficientNet architecture on CIFAR-100 dataset. We also evaluate the Transfer Learning performance of the models trained on ImageNet. The $k$ hyperparameter related to the motivation condition is fine-tuned using 5-fold cross-validation on the training set.

For all our experiments, Arch-X-C refers to the accuracy of a model \textit{Arch-X} trained classically where X denotes its position in the scalable architecture \textit{Arch}. Arch-X-Y-B refers to the accuracy of the model \textit{Arch-X} when it is trained using our method as a base model with \textit{Arch-Y} as a motivated model. Y denotes a higher position in the scalable architecture than X. In turn, Arch-X-Y-M denotes the accuracy of the model \textit{Arch-Y} when it is used as a motivated model during a motivation-inspired training with \textit{Arch-X} as a base model.

In order to demonstrate the efficiency of our method, we use an accuracy per FLOPs metric (ACC/FLOPs) and its associated ratio (ACC/F\_Ratio). Specifically, since we aim for an intermediary performance between the classical training of a base model (Arch-X-C) and the classical training of the motivated model (Arch-Y-C), we compare the gains between using our motivated training to train \textit{Arch-X} (Arch-X-Y-B) versus the classically trained motivated model (Arch-Y-C). In terms of efficiency, we aim to evaluate the cost of replacing Arch-X-C with Arch-X-Y-B versus replacing it with Arch-Y-C. The metric ACC/FLOPs gives the accuracy gained per added FLOPs computation when replacing Arch-X-C with Arch-X-Y-B versus replacing it with Arch-Y-C. Let $Acc_{X-Y-B}$ and $F_{X-Y}$ respectively denote the accuracy and FLOPs count of model Arch-X-Y-B, $Acc_{X-C}$ and $F_{X-C}$ respectively denote the accuracy and FLOPs count of model Arch-X-C and $Acc_{Y-C}$ and $F_{Y-C}$ respectively denote the accuracy and FLOPs count of model Arch-Y-C. We compute ACC/FLOPs for Arch-X-Y-B as :

\begin{equation}
  ACC/FLOPs_{X-Y-B}=\frac{Acc_{X-Y-B}-Acc_{X-C}}{F_{X-Y}-F_{X-C}}
  \label{eq:met1}
\end{equation}

We compute ACC/FLOPs for Arch-Y-C as :

\begin{equation}
  ACC/FLOPs_{Y-C}=\frac{Acc_{Y-C}-Acc_{X-C}}{F_{Y-C}-F_{X-C}}
  \label{eq:met2}
\end{equation}

The ratio between the two metrics readily compares the efficiency of the two models and is given as:

\begin{equation}
  ACC/F\_Ratio_{X-Y}=\frac{ACC/FLOPs_{X-Y-B}}{ACC/FLOPs_{Y-C}}
  \label{eq:ratio}
\end{equation}

Moreover, for our training recipe, we compute total FLOPs count by summing the FLOPs cost of forward passes of the base model during non-motivated states along with the FLOPs cost of forward passes of the motivated model during motivated states. For each epoch, motivated states count is determined by the average number of motivated model activations across runs. We finally report the average FLOPs cost of a forward pass: $F_{X-Y}$ by dividing the sum by the total number of forward passes. For the exact values of $F_{X-Y}$ and $F_{X-C}$ of all our experiments, refer to \Cref{sec:supp_flops} in the supplementary material.

\subsection{ResNet}
We use the CIFAR variants of ResNet as described in \cite{he2016deep}. Following \cite{gomes2024adafisher} we train the networks using SGD~\cite{bottou2010large} with 0.9 momentum, 128 batch size and 5e-4 weight decay and stop at 200 epochs. A cosine decay schedule \cite{loshchilov2016sgdr} is applied on the learning rate.

\Cref{tab:resnet_cifar_results} shows evaluation accuracy over 3 runs for base models trained classically (C suffix) and through motivated states. We manage to improve accuracy for all networks on both CIFAR-10 and CIFAR-100. The base model in the context of our training recipe also achieves better efficiency by gaining more accuracy per FLOPs compared to the next-level model (up to 122x more efficient). This is true for all cases but one as shown in bold.

\begin{table*}[t]
\caption{Classification accuracy (\%) $\pm$ standard deviation and efficiency metrics (ACC/FLOPs with ACC/F\_Ratio between parentheses) on CIFAR-10 and CIFAR-100 for motivation-enhanced ResNet (5 runs) and ViT (3 runs) architectures and baselines.}
\centering
\begin{tabular}{lcccc}
\hline
Model            & \multicolumn{1}{c}{\textbf{CIFAR-10 (\%)}} & \textbf{ACC/FLOPs ($10^{-2}$)} & \multicolumn{1}{c}{\textbf{CIFAR-100 (\%)}} & \textbf{ACC/FLOPs ($10^{-2}$)} \\ \hline
Resnet-20-C      & 92.62 $\pm$ 0.17                  & /                 & 68.86 $\pm$ 0.23                   & /                 \\ \cline{3-3} \cline{5-5} 
Res-20-32B       & \textbf{92.83 $\pm$ 0.20}                  & \textbf{2.18 (1.24)}              & \textbf{69.00 $\pm$ 0.48}                   & \textbf{6.90 (2.23)}              \\ \cline{1-2} \cline{4-4}
Resnet-32-C      & 93.63 $\pm$ 0.13                  & 1.76              & 70.63 $\pm$ 0.38                   & 3.09              \\ \cline{3-3} \cline{5-5} 
Res-32-44B       & \textbf{93.66 $\pm$ 0.05}                  & 0.31 (0.60)           & \textbf{70.67 $\pm$ 0.41}                   & \textbf{1.73 (1.24)}              \\ \cline{1-2} \cline{4-4}
Resnet-44-C      & 93.93 $\pm$ 0.12                  & \textbf{0.52}              & 71.43 $\pm$ 0.23                   & 1.39              \\ \cline{3-3} \cline{5-5} 
Res-44-56B       & \textbf{94.10 $\pm$ 0.19}                  & \textbf{40.48 (122.67)}             & \textbf{71.76 $\pm$ 0.42}                   & \textbf{76.74 (41.04)}             \\ \cline{1-2} \cline{4-4}
Resnet-56-C      & 94.12 $\pm$ 0.23                  & 0.33              & 72.50 $\pm$ 0.28                   & 1.87              \\ \hline\hline
ViT-tiny-C       & 89.70  $\pm$ 0.12                 & /                 & 70.38 $\pm$ 0.68                   & /                 \\ \cline{3-3} \cline{5-5} 
ViT-tiny-small-B & \textbf{89.74 $\pm$ 0.25}                  & \textbf{173.91 (1.75)}            & \textbf{70.53 $\pm$ 0.44}                   & \textbf{3750 (84.19)}              \\ \cline{1-2} \cline{4-4}
ViT-small-C      & 92.98  $\pm$ 0.14                 & 99.39             & 71.85 $\pm$ 0.55                   & 44.54             \\ \hline
\end{tabular}
\label{tab:resnet_cifar_results}
\end{table*}

\begin{table}[t]
\caption{Classification accuracy (\%) and efficiency metrics (ACC/FLOPs with ACC/F\_Ratio between parentheses) on ImageNet for motivation-enhanced ResNet architecture and baselines.}
\centering
\begin{tabular}{lcc}
\hline
Model        & \multicolumn{1}{c}{\textbf{ImageNet (\%)}} & \textbf{ACC/FLOPs} \\ \hline
Resnet-50-C  & 78.06~\cite{wightman2021resnet}                             & /         \\ \cline{3-3} 
Res-50-101B  & \textbf{78.40}                             & \textbf{8.5 (18.48)}       \\ \cline{1-2}
Resnet-101-C & 79.80~\cite{wightman2021resnet}                             & 0.46      \\ \hline
\end{tabular}
\label{tab:resnet_imagenet_results}
\end{table}

\begin{table*}[h]
\caption{Classification accuracy (\%) $\pm$ standard deviation (3 runs) on Transfer Learning tasks for a motivation-enhanced ResNet-50 model.}
\centering
\begin{tabular}{lccccc}
\hline
Model        & ImageNet (\%) & CIFAR-10 (\%)    & CIFAR-100 (\%)   & Flowers (\%) & Pets (\%) \\ \hline
Res-50-Unmot & 78.06         & 93.71 $\pm$ 0.04 & 74.16 $\pm$ 0.36 &   73.56 $\pm$ 0.66        & 93.48 $\pm$ 0.07              \\
\textbf{Res-50-Mot}   & \textbf{78.40} & \textbf{97.15 $\pm$ 0.03} & \textbf{85.15 $\pm$ 0.16} &  \textbf{94.67 $\pm$ 0.28} & \textbf{96.87 $\pm$ 0.07} \\ \hline
\end{tabular}
\label{tab:transfer_results}
\end{table*}

We also evaluate our method on ImageNet. Our motivation-inspired training is based on the A3 procedure from \cite{wightman2021resnet}. Among other settings, we train for 100 epochs using LAMB~\cite{you2019large} optimizer with 2048 batch size, 0.02 weight decay and 8e-3 base learning rate. For data augmentation, Mixup~\cite{zhang2017mixup}, CutMix~\cite{yun2019cutmix} and RandAugment~\cite{cubuk2020randaugment} are applied. \Cref{tab:resnet_imagenet_results} shows that our motivation-inspired training can be applied to large-scale datasets as well and that it introduces performance gains while being 18x more efficient than the next-level model.

Moreover, we investigate how the performance gains on ImageNet translate to downstream tasks during Transfer Learning. Specifically, we finetune a Resnet-50 model using the final weights of Res-50-C (Res-50-Unmot) and Res-50-101-B (Res-50-Mot) on CIFAR-10, CIFAR-100, Flowers~\cite{nilsback2008automated} and Pets~\cite{parkhi2012cats}. We apply a procedure inspired from \cite{touvron2021training} and \cite{wightman2021resnet}. For all tasks we use SGD with 1e-4 weight decay, 384 batch size and 0.01 learning rate decayed along a cosine schedule. As data augmentation, we use Mixup, CutMix, RandAugment and Repeated Augmentation~\cite{hoffer2020augment,berman2019multigrain}. We fine-tune for 300 epochs for CIFAR datasets, 2000 epochs for Flowers and 1000 epochs for Pets.

\Cref{tab:transfer_results} shows that Transfer Learning tasks accuracies increased by 4\% to 29\% when using motivation-enhanced weights. The increase is most substantial for CIFAR-100 and Flowers. This indicates that representations learned through motivation generalize better and the learned embedding space is semantically richer and more task-agnostic.

\subsection{ViT}
Vision Transformers require a large amount of data making them ill-suited for small-scale datasets like CIFAR-10 and CIFAR-100. In our experiments, however, we do not aim to achieve state-of-the-art performances but rather to demonstrate how motivation-inspired training improves model performances. As such, we choose to train from scratch on CIFAR datasets proving the adaptability of our framework to a diverse range of architectures. 

Following \cite{zhang2025depth}, we train for 300 epochs including 20 warm-up epochs. We use AdamW~\cite{loshchilov2017decoupled} optimizer with 0.05 weight decay, 64 batch size and 2.5e-4 learning rate. CIFAR images are resized to 224x224 and the networks patch size is set to 16. As data augmentation, we use RandAugment, Random Erasing~\cite{zhong2020random}, Mixup and CutMix.

Results are given in \cref{tab:resnet_cifar_results}. We report metrics over 3 runs and demonstrate better performance with our method. The motivation-aware training is also up to 84x more efficient.

\subsection{EfficientNet}
\begin{table}[t]
\caption{Classification accuracy (\%) $\pm$ standard deviation (3 runs) and efficiency metrics (ACC/FLOPs with ACC/F\_Ratio between parentheses) on CIFAR-100 for motivation-enhanced EfficientNet architecture and baselines.}
\centering
\begin{tabular}{lcc}
\hline
\textbf{Model}             & \textbf{CIFAR-100 (\%)}   & \textbf{ACC/FLOPs} \\ \hline
EfficientNet-B0-C & 79.09 $\pm$ 0.50 & /         \\ \cline{3-3} 
Eff-0-1B          & \textbf{79.23 $\pm$ 0.34} & \textbf{2.8 (1.26)}       \\ \cline{2-2}
EfficientNet-B1-C & 79.92 $\pm$ 0.60 & 2.22      \\ \cline{3-3} 
Eff-1-2B          & \textbf{80.37 $\pm$ 0.08} & \textbf{22.5 (6.76)}      \\ \cline{2-2}
EfficientNet-B2-C & 80.18 $\pm$ 0.57 & 3.33      \\ \cline{3-3} 
Eff-2-3B          & \textbf{80.32 $\pm$ 0.24} & \textbf{14 (14.89)}        \\ \cline{2-2}
EfficientNet-B3-C & \textbf{80.93 $\pm$ 0.46} & 0.94      \\ \cline{3-3} 
Eff-3-4B          & 80.81 $\pm$ 0.50 & -6 (-100)       \\ \cline{2-2}
EfficientNet-B4-C & 81.08 $\pm$ 0.20 & \textbf{0.06}      \\ \cline{3-3} 
Eff-4-5B          & \textbf{81.27 $\pm$ 0.27} & \textbf{0.83 (4.61)}      \\ \cline{2-2}
EfficientNet-B5-C & 82.10 $\pm$ 0.33 & 0.18      \\ \hline
\end{tabular}
\label{tab:efficientnet_cifar100_results}
\end{table}

\begin{table}[t]
\caption{Classification accuracy (\%) $\pm$ standard deviation (3 runs) and FLOPs count on CIFAR-100 for motivation-enhanced EfficientNet networks focusing on motivated models along with baselines.}
\centering
\begin{tabular}{l c c}
    \hline
    \textbf{Model} & \textbf{CIFAR-100 (\%)} & \textbf{\#FLOPs} \\
    \hline
    EfficientNet-B0-C    & 79.09 $\pm$ 0.50 & 0.39G\\
    \hline
    Eff-0-1B           & 79.23 $\pm$ 0.34 & 0.44G \\
    \hdashline
    \textbf{Eff-0-1M}        & \textbf{80.09 $\pm$ 0.27} & \textbf{0.44G}\\
    EfficientNet-B1-C    & 79.92 $\pm$ 0.60 & 0.70G \\
    \hline
    Eff-1-2B           & 80.37 $\pm$ 0.08 & 0.75G \\
    \hdashline
    \textbf{Eff-1-2M}        & \textbf{81.02 $\pm$ 0.18} & \textbf{0.75G} \\
    EfficientNet-B2-C    & 80.18 $\pm$ 0.57 & 1.0G \\
    \hline
    Eff-2-3B           & 80.17 $\pm$ 0.23 & 1.13G \\
    \hdashline
    \textbf{Eff-2-3M}    & \textbf{81.32 $\pm$ 0.17} & \textbf{1.13G} \\
    EfficientNet-B3-C    & 80.93 $\pm$ 0.46 & 1.8G \\
    \hline
    Eff-3-4B           & 80.28 $\pm$ 0.46 & 2.18G \\
    \hdashline
    \textbf{Eff-3-4M}           & \textbf{81.60 $\pm$ 0.34} & \textbf{2.18G} \\
    EfficientNet-B4-C    & 81.08 $\pm$ 0.20 & 4.2G \\
    \hline
    Eff-4-5B           & 81.04 $\pm$ 0.40 & 5.14G \\
    \hdashline
    Eff-4-5M           & 81.94 $\pm$ 0.26 & 5.14G \\
    \textbf{EfficientNet-B5-C}    & \textbf{82.10 $\pm$ 0.33} & \textbf{9.9G}\\
    \hline
  \end{tabular}
\label{tab:efficientnet_reg}
\end{table}

We train the EfficientNet networks on CIFAR-100. We do not evaluate past EfficientNet-B5 because the two larger networks no longer improve performance. Following \cite{pittorino2021entropic} and \cite{tan2019efficientnet}, we train for 350 epochs using RMSprop~\cite{hinton2012rmsprop} optimizer with 0.9 momentum, 64 batch size, 10e-5 weight decay and an initial learning rate of 0.01 that decays by 0.97 every 2 epochs after 5 epochs warm-up. We apply Dropout and Stochastic Depth following the values described in \cite{tan2019efficientnet}. To accommodate the networks architecture, images are rescaled to 224x224 and standard preprocessing is applied (horizontal flip and cropping). As data augmentation, we use AutoAugment with CIFAR-10 policy \cite{cubuk2018autoaugment}.

\Cref{tab:efficientnet_cifar100_results} reports evaluation accuracies comparing models trained classically (C suffix) versus using motivation. Our recipe outperforms the classical training in four out of the five evaluated stages and sometimes outperforms the next-level model. It also does so with up to 14x more efficiency.

In the case of the EfficientNet architecture, our method also acted as a regularization mechanism for the motivated model. For hyperparameter $k=2$, we observed that the accuracies of the motivated models were better than when they were trained classically.

Results in \cref{tab:efficientnet_reg} show how the motivated model which was only trained during the motivated states, outperforms the same network trained classically. Not only that, some networks trained as motivated models surpassed bigger networks trained classically. In particular, the regularized B2 network surpassed the classical B2 (33\% more FLOPs), classical B3 (30\% more parameters and 2.4x more FLOPs) and even comes close to classical B4 (double the number of parameters and 5.6x more FLOPs). The regularized B3 network also outperforms classical B3 (59\% more FLOPs) and classical B4 (58\% more parameters and 3.7x more FLOPs). As stated in \cref{sec:relwork}, our recipe drops the differential layers during normal states and activates them during motivated states similarly to how Dropout regularizes networks and improves generalization.

Moreover, since the base model offers comparable performance to when it is trained classically, a train once, deploy twice procedure is possible. Namely, it is possible to train both a base model and a motivated one at the same time while keeping the accuracy of the former and improving the accuracy of the latter. The cost of training both is also lower than training the bigger model in terms of FLOPs since the forward and backward operations include only a subset of the motivated model during non motivated states.

\begin{table*}
\caption{Classification accuracy (\%) $\pm$ standard deviation on Resnet/CIFAR-10 (5 runs) and EfficientNet/CIFAR-100 (3 runs) exploring the impact of the motivation condition.}
\label{tab:abla}
\centering
\begin{tabular}{ll ll}
    \hline
    \textbf{Resnet} & \textbf{CIFAR-10 (\%)} & \textbf{Efficientnet} & \textbf{CIFAR-100 (\%)} \\
    \hline
    Resnet-20-C             & 92.62 $\pm$ 0.17 & Efficientnet-B0-C         & 79.09 $\pm$ 0.50 \\
   \textbf{Res-20-32-B}             & \textbf{92.83 $\pm$ 0.20} & \textbf{Eff-0-1-B }                 & \textbf{79.23 $\pm$ 0.34} \\
    Res-20-32-B-EXP-A       & 92.38 $\pm$ 0.32 & Eff-0-1-B-EXP-A            & 79.01 $\pm$ 0.44 \\
    Res-20-32-B-EXP-B       & 92.62 $\pm$ 0.36 & Eff-0-1-B-EXP-B            & 79.13 $\pm$ 0.19 \\
    \hline
    Resnet-32-C             & 93.63 $\pm$ 0.20       & Efficientnet-B1-C          & 79.92 $\pm$ 0.60 \\
    \hline
  \end{tabular}
\end{table*}
\subsection{Ablation study}
\subsubsection{Relevance of the motivation condition}
We conduct an ablation study to assess the impact of the motivation condition. We design two experiments.
\begin{itemize}
    \item \textbf{Experiment A:} for each epoch, we randomly determine a number of times in which we activate the motivated model. Then, we randomly generate the exact batch indices for the activation.
    \item \textbf{Experiment B:} we use the previous experiments to calculate the average number of times (across runs) the motivated model is activated at each specific epoch using the motivated condition outlined in \cref{sec:motivcond}. However, the exact batch indices at which the motivated model is activated are generated randomly.
\end{itemize}
We conduct evaluations on ResNet for CIFAR-10 and EfficientNet for CIFAR-100. We keep the training configuration the same as in the original experiments. The results reported in \cref{tab:abla} show that Experiment A deteriorates performance compared to the classical training of the base model. This means that training the base model as a part of a bigger model at completely random times hurts its performance. We expected this on an intuitive level because, ultimately, we are disrupting the training and using weights from the motivated model in the base one whereas these same weights expect other layers down the line in the forward pass that are not present in the base model. However, surprisingly, activating the motivated model a specific number of time, an information leaked from the motivation condition, but with no condition on the exact batch indices did not hurt performance for Resnet and even achieved an increase in performance for EfficientNet. The activation of the motivated model exactly at the motivation condition provided the best performance as shown in bold proving its importance but Experiment B hints at the relative importance of the number of these activations independently of their moment of occurrence during training.

\subsubsection{Other motivation conditions}
Drawing from human experience, motivation notably occurs when learning is met with perceived success. In the context of artificial neural networks, one can imagine many motivation conditions. We explore in this section alternate instantiations of this component within our framework.

\begin{itemize}
    \item \textbf{EMA on Training Loss:} We employ an EMA on the training loss in order to accommodate mini-batch noise. We set the smoothing factor $\alpha = 0.02$ and consider the loss has decreased if it drops by 0.2\%. These parameters were empirically determined to allow for a sufficient number of motivated model activations comparable with that of the original experiments. We activate the larger model after $k$ consecutive decreases. We refer to this experiment with the \textbf{ema} suffix.
    \item \textbf{Validation Loss:} Instead of activating the bigger network when training loss decreases for k consecutive batches, we use validation loss. Running a full forward pass on a validation dataset for each batch in the training set is costly. To maintain reasonable execution times, we spare only one batch from the training set and use it for validation. We refer to this condition using the \textbf{valloss} suffix.
    \item \textbf{Gradient Slope:} A training that is going well can manifest as an increasing gradient magnitude signalling a steeper region in the training loss. We activate the motivated model if the euclidean norm of the concatenated gradients vector increases for $k$ consecutive batches. We use the \textbf{slope} suffix for this experiment.
\end{itemize}

We rerun an experiment on Resnet for CIFAR-10 changing only the motivation condition. $k$ is fine-tuned using 5-fold cross-validation on the training set. \Cref{tab:oth_motiv} shows the superiority of the condition we adopted in our framework. The other conditions are all also effective at improving the base model at varying degrees.

\begin{table}
\caption{Classification accuracy (\%) $\pm$ standard deviation (5 runs) of a Resnet architecture on CIFAR-10 using alternate motivation conditions.}
\centering
\begin{tabular}{l c}
    \hline
    \textbf{Model} & \textbf{CIFAR-10 (\%)} \\
    \hline
    Resnet-20-C             & 92.62 $\pm$ 0.17\\
    \textbf{Res-20-32-B-original}    & \textbf{92.83 $\pm$ 0.20} \\
    Res-20-32-B-ema           & 92.78 $\pm$ 0.23 \\
    Res-20-32-B-valloss           & 92.65 $\pm$ 0.17 \\
    Res-20-32-B-slope           & 92.74 $\pm$ 0.24 \\
    \hline
  \end{tabular}
  \label{tab:oth_motiv}
\end{table}

\section{Conclusion}
In this work, we take inspiration from affective neuroscience and propose an alternating training scheme with conditional capacity expansion that mimics high states of curiosity in the human brain. It trains a base model continuously and intermittently activates a larger motivated model during a predefined motivation condition (loss decreasing for k consecutive batches). Extensive experiments on multiple scalable architectures (ResNet, ViT and EfficientNet) and datasets (CIFAR, Imagenet, Flowers and Pets), demonstrate an efficiently improved accuracy and generalization in the base model and in the case of EfficientNet, improved accuracy of the motivated model over its standalone version despite being inactive for certain periods during the training. Then, our method allows for a "train once, deploy twice" scheme producing two high-performing models with distinct computational footprints which is appealing for resource-constrained configurations. Future research will focus on exploring more sophisticated, learnable and task-driven motivation conditions, formalizing the theoretical underpinnings of motivation-based switching, including why the motivated model is a high-performer in the EfficientNet architecture only and investigating less heuristical weight mapping such as mapping layers that exhibit high functional similarity. We also plan to get closer to the biological inspiration for our work where training and inference happen simultaneously and extend our framework to the online learning setting.
\clearpage
{
    \small
    \bibliographystyle{ieeenat_fullname}
    \bibliography{main}
}

\clearpage
\setcounter{page}{1}
\maketitlesupplementary

\section{FLOPs costs of the models}
\label{sec:supp_flops}
The FLOPs cost of a model is traditionally given as the cost of one forward pass for a single image. As such, it is influenced by the model architecture and the size of the input image. It is the one referred to in \Cref{sec:exp} by $F_{X-C}$.

In traditional training, the model is kept the same throughout and the inference FLOPs count is a valid metric to compare training cost across models. In our case however, the model changes during training. One solution is to compute the average cost of a forward pass during our dual-training procedure. We start by computing the total FLOPs count. We sum the FLOPs cost of forward passes of the base model during non-motivated states along with the FLOPs cost of forward passes of the motivated model during motivated states. For each epoch, motivated states count is determined by the average number of motivated model activations across runs. The average FLOPs cost of a forward pass $F_{X-Y}$ referenced in \Cref{sec:exp} is then given by dividing the sum by the total number of forward passes. It is to be noted that $F_{X-Y}$ is the cost of training both base and motivated networks, since they are trained simultaneously.

Since $F_{X-Y}$ depends on the number of motivated model activation, then it is dependent on more parameters than the models architecture and the size of the input image. Considering the motivation condition we instantiated our framework through, any hyperparameter that influences the evolution of the loss function will have an impact on the training cost, even the seeds used. Moreover, different datasets create different loss landscapes and as such will influence the value of $F_{X-Y}$.

Tables \ref{tab:resnet_cifar_FLOPs}, \ref{tab:resnet_imagenet_FLOPs}, \ref{tab:vit_cifar_FLOPs} and \ref{tab:eff_cifar_FLOPs} give the FLOPs count of all our training experiments. They are detailed per architecture and per dataset. As explained above, the classical training FLOPs count is not influenced by the dataset contrary to our dual-training.

\begin{table}[h]
\caption{FLOPs Count on CIFAR-10 and CIFAR-100 for motivation-enhanced ResNet architecture.}
\centering
\begin{tabular}{lcc}
\hline
Model       & CIFAR-10 & CIFAR-100 \\ \hline
Resnet-20-C & 82.82M   & 82.82M    \\
Res-20-32B  & 92.47M   & 85.12M    \\ \hline
Resnet-32-C & 140.13M  & 140.13M   \\
Res-32-44B  & 149.67M  & 142.44M   \\ \hline
Resnet-44-C & 197.44M  & 197.44M   \\
Res-44-56B  & 197.86M  & 197.87M   \\ \hline
Resnet-56-C & 254.76M  & 254.76M   \\ \hline
\end{tabular}
\label{tab:resnet_cifar_FLOPs}
\end{table}

\begin{table}[h]
\caption{FLOPs Count on ImageNet for motivation-enhanced Resnet architecture.}
\centering
\begin{tabular}{lc}
\hline
Model        & ImageNet \\ \hline
Resnet-50-C  & 3.8G     \\
Res-50-101-B & 3.84G    \\ \hline
Resnet-101-C & 7.6G     \\ \hline
\end{tabular}
\label{tab:resnet_imagenet_FLOPs}
\end{table}

\begin{table}[h]
\caption{FLOPs Count on CIFAR-10 and CIFAR-100 for motivation-enhanced ViT architecture.}
\centering
\begin{tabular}{lcc}
\hline
Model            & CIFAR-10 & CIFAR-100 \\ \hline
ViT-tiny-C       & 1.3G     & 1.3G      \\
ViT-tiny-small-B & 1.323G   & 1.304G    \\ \hline
ViT-small-C      & 4.6G     & 4.6G      \\ \hline
\end{tabular}
\label{tab:vit_cifar_FLOPs}
\end{table}

\begin{table}[h]
\caption{FLOPs Count on CIFAR-100 for motivation-enhanced EfficientNet architecture.}
\centering
\begin{tabular}{lc}
\hline
Model             & CIFAR-100 \\ \hline
EfficientNet-B0-C & 0.39G     \\
Eff-0-1B          & 0.44G     \\ \cline{2-2} 
EfficientNet-B1-C & 0.7G      \\
Eff-1-2B          & 0.72G     \\ \cline{2-2} 
EfficientNet-B2-C & 1G        \\
Eff-2-3B          & 1.01G     \\ \cline{2-2} 
EfficientNet-B3-C & 1.8G      \\
Eff-3-4B          & 1.82G     \\ \cline{2-2} 
EfficientNet-B4-C & 4.2G      \\
Eff-4-5B          & 4.43G     \\ \cline{2-2} 
EfficientNet-B5-C & 9.9G      \\ \hline
\end{tabular}
\label{tab:eff_cifar_FLOPs}
\end{table}

\begin{figure*}
    \centering
    {\small
    \begin{tabular}{cc}
        \includegraphics[width=0.48\linewidth]{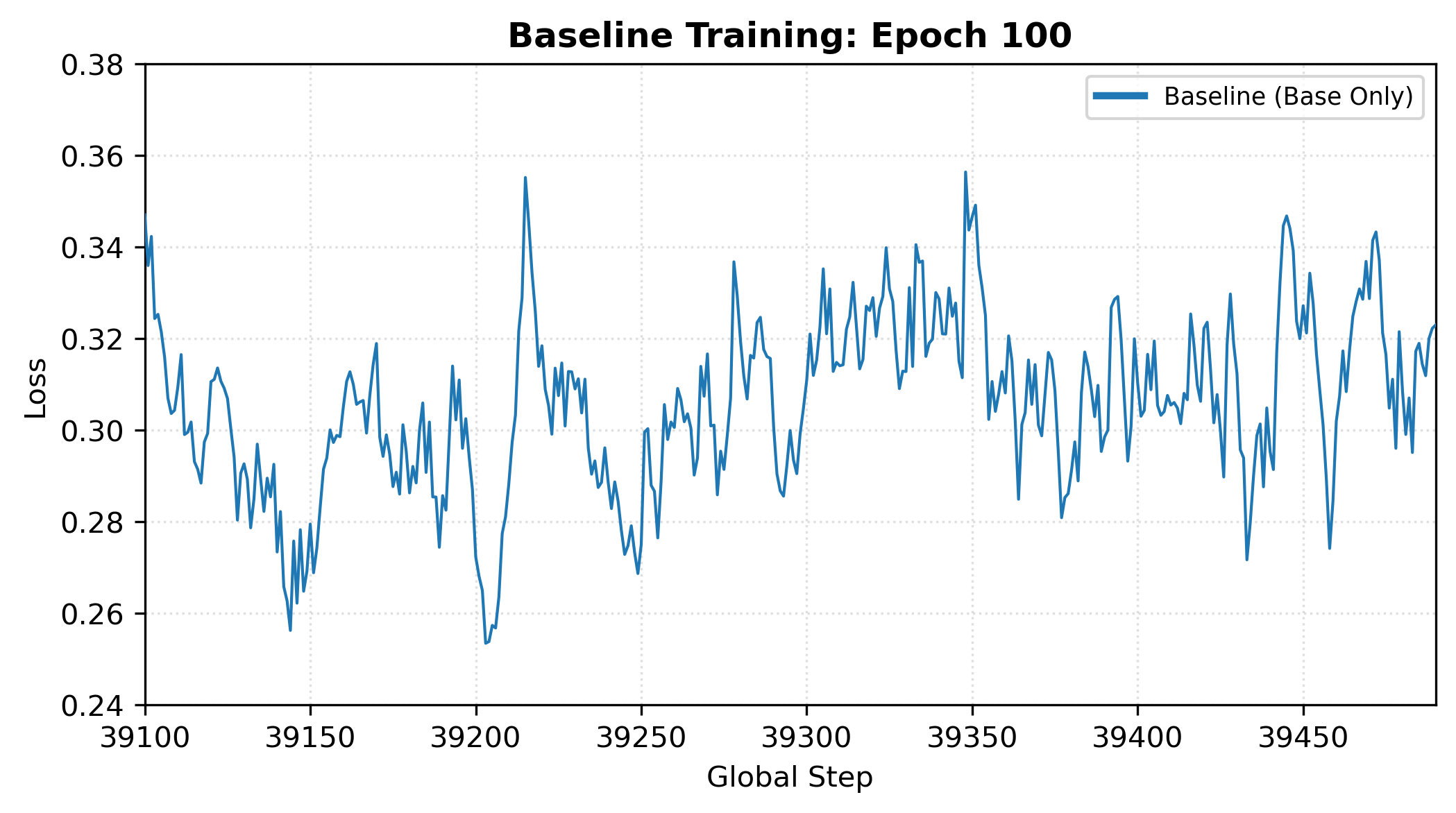} &
        \includegraphics[width=0.48\linewidth]{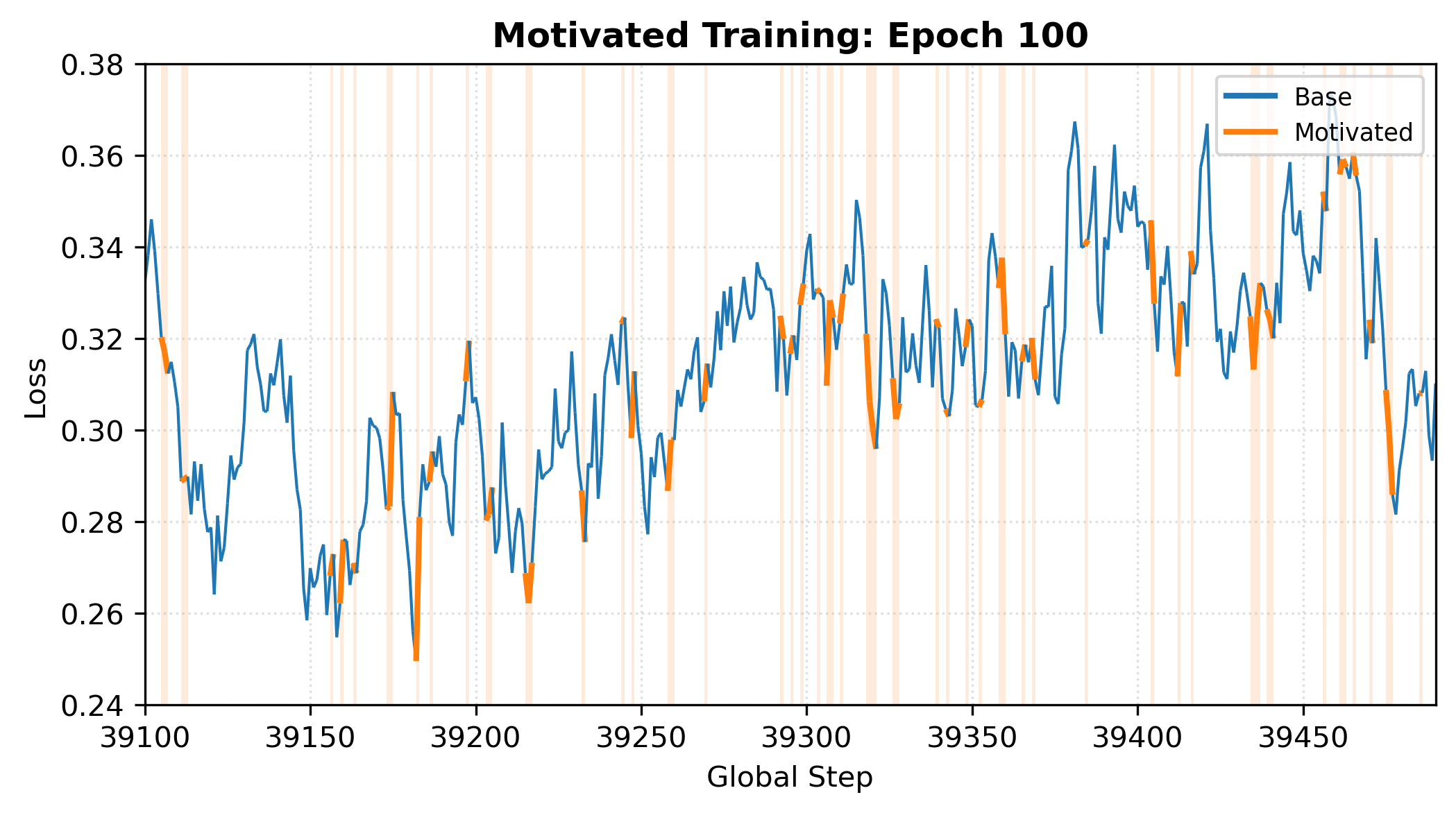} \\
        \includegraphics[width=0.48\linewidth]{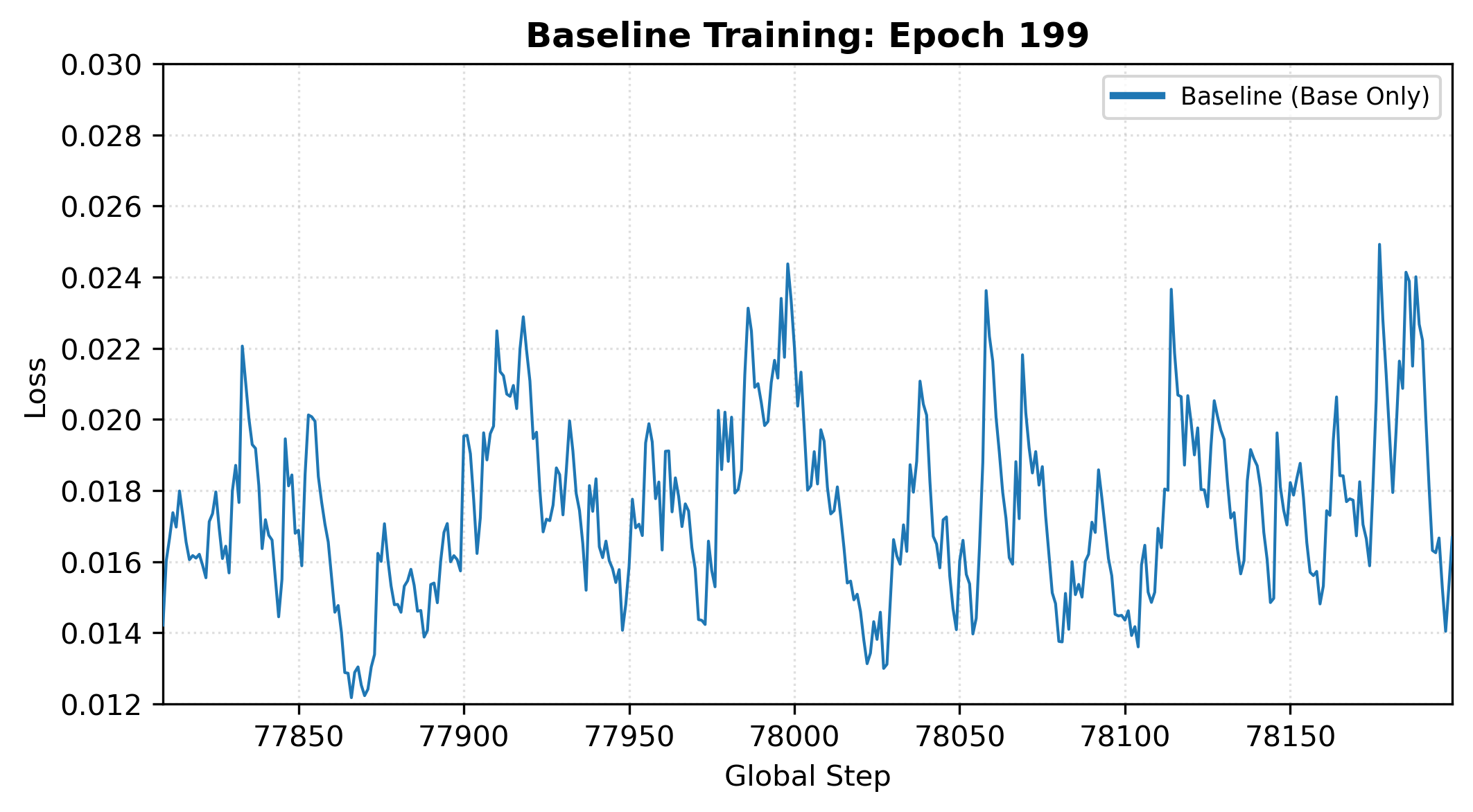} &
        \includegraphics[width=0.48\linewidth]{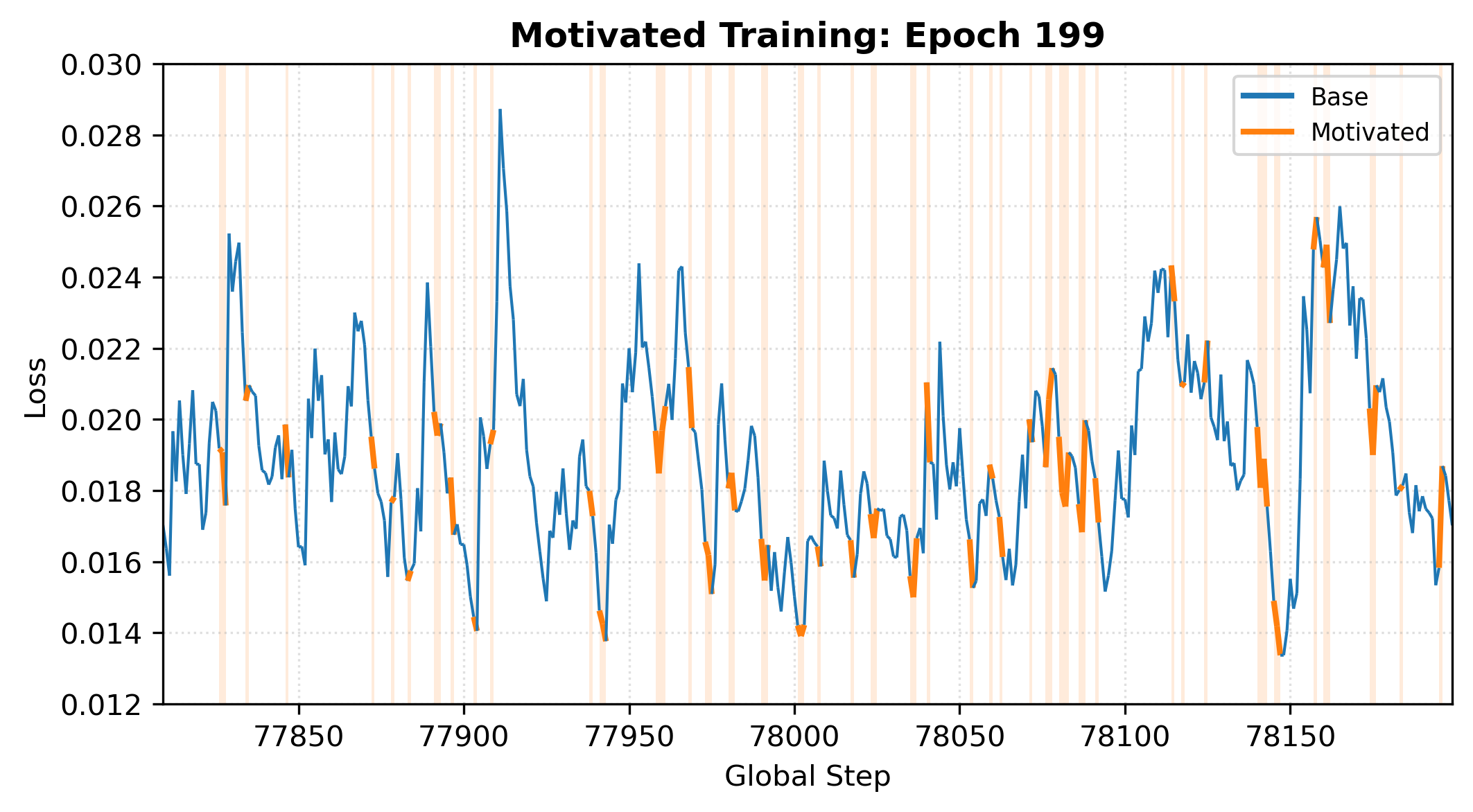} \\
    \end{tabular}
    \captionof{figure}{Training dynamics under cosine learning rate for Resnet-20-C (baseline) and Res-20-32B (motivated training).
    }
    \label{fig:training_dynamics_rebuttal}
    }
\end{figure*}

\section{Training Dynamics}
In this section we explore the impact of our method on training dynamics. We include Fig. \ref{fig:training_dynamics_rebuttal} comparing baseline (Resnet-20-C) and motivation-based training (Res-20-32B) at epochs 100 and 199 (last epoch) under identical axis scales. After 200 epochs on CIFAR-10, motivation-based training achieves 92.91\% accuracy versus 92.58\% for the baseline. While both methods occasionally reach similar instantaneous loss values, motivation-based switching primarily alters late-stage optimization dynamics. Motivated steps frequently align with directional transitions in the loss trajectory, including rapid descent after plateaus and ascent after local minima, suggesting that intermittent capacity expansion perturbs the local optimization geometry rather than enforcing monotonic loss reduction. These transitions reduce loss variance, bias optimization toward flatter regions of the loss landscape, and help explain why the heuristic loss-based trigger is effective in practice, offering a plausible mechanism for the observed generalization and transfer gains.

\end{document}